%% file: main.tex
\documentclass[10pt,twocolumn,letterpaper]{article}

\usepackage{cvpr}      
\input{preamble}
\definecolor{cvprblue}{rgb}{0.21,0.49,0.74}
\usepackage[pagebackref,breaklinks,colorlinks,allcolors=cvprblue]{hyperref}

\def\confName{CVPR}
\def\confYear{2027}

\title{A Study on Performance Enhancement of a Webcam-Based Real-Time Eye-Tracking System}

\author{Minuk Jung\\
Kookmin University\\
Seoul, Republic of Korea\\
{\tt\small alsdnr5000@kookmin.ac.kr}
\and
Kyeonghun KIM\\
OUTTA\\
Seoul, Republic of Korea\\
{\tt\small kyeonghun.kim@outta.ai}
\and
Sooyeon Yoon\textsuperscript{\dag}\\
Kookmin University\\
Seoul, Republic of Korea\\
{\tt\small 1104py@kookmin.ac.kr}
}

\usepackage{float}
\usepackage{placeins}
\begin{document}
\maketitle

\begingroup
\renewcommand{\thefootnote}{\dag}
\footnotetext[1]{Corresponding author: Sooyeon Yoon
(1104py@kookmin.ac.kr)}
\endgroup

\input{sec/0_abstract}    
\input{sec/1_intro}
\input{sec/2_formatting}
\input{sec/3_finalcopy}
\input{sec/4_Experiment}
\input{sec/5_conclusion}

\nocite{*}
{
    \small
    \bibliographystyle{ieeenat_fullname}
    \bibliography{main}
}


\end{document}

%% file: sec/0_abstract.tex
\begin{abstract}
This study proposes a real-time gaze-tracking system that uses a standard RGB webcam to estimate screen coordinates. The system combines ETH-XGaze-based gaze estimation with nine-point calibration and polynomial ridge regression. To reduce errors caused by changes in head position, it also applies a conditional residual correction derived from head and pose features extracted with RTMPose. Whether this correction is used depends on the reliability of the detected features and the characteristics of the calibration data. In a preliminary experiment, one participant completed trials at 12 screen targets. The resulting mean screen-coordinate error was 204.24 pixels, which was 35.83\% lower than that of xgaze-js and 57.69\% lower than that of gaze\_tracker. The system produced valid gaze estimates for all 12 targets. These preliminary findings demonstrate the feasibility of combining user-specific calibration with conditional pose-based correction for webcam gaze tracking. Because the evaluation involved only one participant, further experiments with more participants and a wider range of postures are required.
\end{abstract}

%% file: sec/1_intro.tex
\section{Introduction}
\label{sec:intro}
Eye tracking has been studied for decades as a means of understanding eye movements and visual attention. With the development of computer vision and deep learning, its use has extended beyond laboratory settings to various human-computer interaction (HCI) applications. Examples include assistive technologies for people who have difficulty using their hands or speech to communicate, educational systems that examine students' visual attention and screen exploration behavior, and driver monitoring systems (DMS) designed to identify gaze deviation and distraction [1]. Although these applications serve different purposes, they share a practical requirement: gaze information must remain sufficiently accurate and available during use. Conventional eye tracking systems often rely on dedicated optical devices or infrared (IR) sensors [2]. Such equipment can provide precise measurements, but its cost and installation requirements may limit its use in everyday environments. Standard RGB cameras offer a more accessible alternative, and several studies have investigated deep learning methods for estimating gaze from ordinary camera images. The challenge is that webcam images are affected by factors that are difficult to control outside a laboratory. Differences in facial and eye characteristics, lighting, camera placement, viewing distance, and head orientation can all influence gaze estimation results [3]. Estimating gaze direction alone is not sufficient when the application requires a position on a computer screen. The estimated direction must also be mapped to two-dimensional screen coordinates. This mapping is commonly established through user-specific calibration, in which a user looks at several known screen locations while the system records the corresponding gaze features. A mapping model is then fitted to these observations. Calibration can account for individual differences and the display setup, but its effectiveness depends partly on how closely the conditions during use resemble those during calibration. For example, if a user leans forward or turns their head after calibration, the relationship between the recorded gaze features and screen positions may change. The original mapping may no longer represent the new viewing condition accurately, leading to larger coordinate errors [4]. This issue suggests that calibration alone may not be sufficient for gaze tracking under changing postures. In this work, we investigate real-time screen gaze tracking using a standard RGB webcam. Rather than developing a new gaze estimation network, we combine an existing gaze estimation method with user-specific screen coordinate mapping and correction based on facial and posture features. We examine the resulting system through a comparison with existing webcam-based gaze tracking methods, focusing on screen coordinate error and the proportion of valid tracking outputs. The experiments provide an initial assessment of the proposed approach and help identify the limitations that remain when gaze tracking is performed with an ordinary webcam.

%% file: sec/2_formatting.tex
\section{Related Work}
\label{sec:formatting}

\subsection{Related Work on Eye Tracking}
 Eye tracking methods can be broadly divided into approaches that use geometric features of the eyes and approaches that estimate gaze direction by learning appearance features from facial images. Depending on their objectives, these methods predict gaze regions, gaze directions, or screen gaze coordinates. Therefore, their evaluation metrics and application scopes should be considered when comparing performance. Real-time Eye and Gaze Position Tracking [5] used infrared illumination to detect the pupil and corneal reflections and classified gaze regions using a General Regression Neural Network (GRNN). The screen was divided into eight regions in a 4 × 2 grid, and classification performance was evaluated under head translation and rotation conditions. The study reported a success rate of approximately 90\% under facial movement conditions and 85\% for users not included in the training data. These results demonstrate the potential for gaze region classification under head movement. However, the method relies on infrared equipment and pupil and corneal reflection detection. Furthermore, its evaluation does not measure continuous screen coordinate errors. Therefore, its classification accuracy cannot be directly compared with pixel-level gaze position accuracy. ETH-XGaze [6] introduced a large-scale facial image dataset and benchmark for gaze estimation, covering a wide range of head poses and gaze directions. The method estimates gaze direction from facial images and evaluates performance using the mean angular error between predicted and ground-truth gaze directions. The study also examined within-dataset evaluation, cross-dataset evaluation, user-specific calibration, and performance variations across head poses and gaze directions. ETH-XGaze provides a foundation for training and evaluating appearance-based gaze estimation under diverse head poses. However, differences in image quality and capture conditions exist between controlled data collection environments and standard webcam settings. Additional screen coordinate mapping is also required to convert estimated gaze directions into gaze positions on a user's monitor. Incorporating shoulder and upper-body posture information into screen coordinate estimation remains an additional research challenge. An AI-Based Video Analysis System for Early Detection of Externalizing Disorders in Children [7] is an application-oriented study that uses gaze direction and behavioral information. The system detects the body and face and determines whether the estimated gaze direction intersects a predefined region of interest (ROI) to classify attentive and inattentive states. The study tested L2CS-Net [8] and VSL2G-Net, a gaze estimation model incorporating video stabilization [9]. Classification accuracy was used to evaluate attention states, while precision, recall, and F1-score were used for separate behavioral analysis. This study demonstrates the potential of integrating gaze information into behavioral analysis. However, attention classification based on whether a user looks at a predefined ROI should be distinguished from screen gaze coordinate accuracy. In addition, looking at a specific region does not necessarily indicate attention. As described above, previous eye tracking studies have different objectives and evaluation criteria, including gaze region classification, gaze direction estimation, and attention state analysis. In this study, we use a pretrained gaze estimation model from the ETH-XGaze family to estimate gaze direction and apply user-specific nine-point calibration data collected using a standard RGB webcam to obtain continuous two-dimensional screen coordinates. Our work focuses on integrating an existing gaze estimation model into a practical screen gaze tracking system. Performance is evaluated primarily using the pixel distance between target screen positions and predicted gaze coordinates, along with screen-size-normalized error.

\subsection{Related Work on Gaze and Pose estimation}
Gaze direction is influenced by eye movements and head orientation, while changes in body posture are also important factors to consider. Previous studies have explored various approaches to predicting gaze direction using head and body posture information or analyzing the coordination of eye, head, and torso movements during gaze shifts. Other studies have also used gaze and posture information to analyze learners' behavior and states. Pose2Gaze [10] is a model that predicts three-dimensional gaze direction using head orientation and full-body pose sequences. It extracts directional features from head orientation and spatial relationships between joints and temporal movement features from full-body poses to predict gaze direction. The model was evaluated on the MoGaze, ADT, GIMO, and EgoBody datasets using the mean angular error between predicted and ground-truth gaze vectors. Ablation studies were also conducted to analyze the contribution of full-body pose information. The results suggest that head orientation and body movement contain useful information for gaze prediction. However, because the model does not directly use facial or eye images, it may have difficulty distinguishing gaze changes caused only by eye movements while body posture remains unchanged. Furthermore, since the model outputs three-dimensional gaze directions, additional screen coordinate mapping is required to estimate gaze positions on an individual user's monitor. Eye, Head and Torso Coordination During Gaze Shifts in Virtual Reality [11] investigated how the eyes, head, and torso move together during gaze shifts in a virtual reality (VR) environment. Using an eye tracker, a head-mounted display (HMD), and torso sensors, the study evaluated movement onset times, target acquisition times, the contributions of individual body parts, and differences between sitting and standing postures. The findings provide evidence that torso movement, in addition to eye and head movements, should be considered when analyzing gaze shifts. However, the study primarily focused on coordination between body parts rather than the accuracy of screen gaze coordinate estimation. Furthermore, additional validation is needed before applying findings from controlled VR experiments to standard webcam and monitor environments. A Study on Improving Learning Efficiency Based on Posture Analysis and Gaze Recognition [12] proposed a system that provides warnings to learners based on posture analysis and gaze direction. For posture analysis, the positions of the person, chair, head, back, and hips were compared with a reference posture. For gaze analysis, the study compared direction classification results using only pupil position with those obtained by additionally considering facial orientation. The results suggest that incorporating facial orientation may improve gaze direction classification. However, the posture experiments and gaze evaluation cases were limited, and the study did not report pixel-level errors for continuous screen coordinates or angular errors for gaze direction. Therefore, it is difficult to assess the system's precision and generalization performance. In addition, posture and gaze were analyzed separately, which differs from approaches that combine both types of information during screen gaze coordinate estimation. These studies demonstrate the relationship between gaze and body posture and the potential benefits of combining them. However, further validation is required for real-time screen gaze tracking systems that integrate user-specific screen mapping and posture information using standard webcams. In this study, we develop an integrated pipeline that combines facial image-based gaze estimation, screen coordinate mapping and error correction through nine-point calibration, posture information extracted using RTMPose[13], and screen coordinate recording and visualization. Posture-based correction is applied based on internal validation and input validity checks. When the required conditions are not satisfied, the system uses the baseline screen coordinates without posture-based correction.

%% file: sec/3_finalcopy.tex
\section{Experimental Design}
\subsection{Eye tracking}
 In this study, we use XGaze-js, a gaze estimation model based on ETH-XGaze, to estimate gaze direction using a standard RGB webcam [14]. We use a pretrained XGaze-js model and apply its estimated gaze features to user-specific calibration and posture-based correction. ETH-XGaze provides a deep learning-based framework for estimating gaze direction from facial images, using training data that cover a wide range of head poses and gaze directions. The gaze estimation process begins with detecting facial landmarks in the input image using MediaPipe [15]. Based on the detected landmarks, the position and orientation of the face are estimated. The facial image is then aligned to a standardized size and orientation before being passed to the gaze estimation model. Next, XGaze-js estimates the user's gaze direction as horizontal and vertical angles. These angles are transformed with respect to the camera coordinate system and represented as two-dimensional gaze features for subsequent screen coordinate mapping.

\subsection{Calibration and Screen Coordinate Mapping}
 In gaze estimation using a standard RGB webcam, differences may occur between the estimated gaze direction and the actual gaze position on the screen due to individual facial and eye characteristics, webcam position, display size, and environmental conditions. To reduce these differences, previous eye tracking studies have used calibration methods that collect gaze data while users look at predefined reference points on the screen and learn the relationship between gaze direction and actual screen coordinates [16].
Based on this approach, we construct a user-specific screen coordinate mapping model using nine-point calibration and second-order polynomial ridge regression [17]. During calibration, nine reference points are arranged in a 3 × 3 grid, with three points each in the upper, middle, and lower regions of the screen. The user is instructed to look at each reference point sequentially. The first second after each point appears is excluded to reduce the effects of gaze transitions. The stability of the subsequently collected gaze features is then evaluated.
Outliers are removed using the Median Absolute Deviation (MAD) method, and representative gaze features are extracted for each reference point. Second-order polynomial ridge regression is then used to model the nonlinear relationship between the extracted gaze features and the corresponding screen coordinates. During this process, the weights of the gaze estimation neural network remain unchanged. Only the user-specific calibration data are used to train the screen coordinate mapping model. The trained mapping model converts gaze features estimated in real time into two-dimensional screen coordinates. This allows the system to account for individual gaze characteristics and differences in webcam and display environments.

\begin{table}[h]
    \centering
    \caption{Calibration task.}
    \label{tab:calibration}

    \setlength{\tabcolsep}{2pt}
    \renewcommand{\arraystretch}{1.1}

    \begin{tabular}{|c|c|}
        \hline
        \includegraphics[
            width=0.45\columnwidth,
            height=2.5cm,
            keepaspectratio
        ]{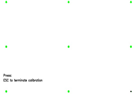}
        &
        \includegraphics[
            width=0.45\columnwidth,
            height=2.5cm,
            keepaspectratio
        ]{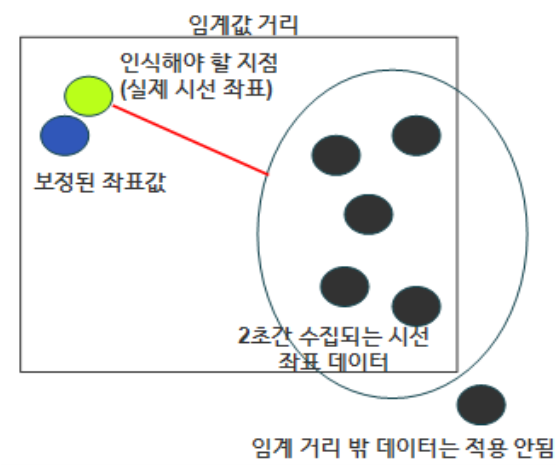}
        \\
        \hline
        (a) Calibration screen
        &
        (b) Calibration workflow
        \\
        \hline
    \end{tabular}
\end{table}

\subsection{RTMPose-Based Posture Feature Extraction}
In standard webcam environments, changes in head position or upper-body posture may cause differences between the posture conditions during initial calibration and actual use. These changes can affect the relationship between gaze direction and screen coordinates. To address this issue, we incorporate RTMPose-based pose estimation into the proposed system. RTMPose is a deep learning-based pose estimation model that detects key body joints and landmarks from input images. In this study, we use RTMPose to these keypoints, we calculate the shoulder center position, distance between the shoulders, relative position of the head and shoulders, an image-based indicator of head tilt, and shoulder line inclination. These measurements are combined to construct a seven-dimensional posture feature vector representing the user's head and upper-body posture.
To reduce sudden fluctuations in keypoint positions during detection, we apply temporal smoothing and verify detection confidence and keypoint validity [18]. Finally, the extracted posture features are combined with gaze estimation results and used for residual correction of screen coordinates.

\begin{figure}[h]
  \centering
  \centering
  \includegraphics[width=\linewidth]{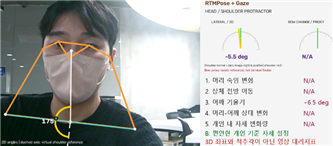}
  \caption{Pose Estimation Using RTMPose}
  \label{fig:rtmposel}
\end{figure}

\subsection{Conditional Residual Correction}
In RGB webcam-based gaze tracking, user-specific calibration can correct the relationship between gaze direction and screen coordinates. However, changes in head position or upper-body posture may cause differences between the initial calibration conditions and actual usage conditions [19]. These changes can affect screen gaze estimation accuracy. To address this issue, we apply a conditional residual correction method that combines gaze estimation results with facial and posture features. Conditional residual correction learns the difference between the baseline screen coordinates predicted through nine-point calibration and the actual target coordinates. The learned residuals are then used to correct screen coordinates during real-time gaze tracking. While the baseline screen mapping model converts gaze features into screen coordinates, the residual correction model aims to reduce additional coordinate errors associated with facial and posture features. In this study, we construct two models: a facial feature-based model that represents face position and head orientation, and a posture-enhanced model that incorporates posture features extracted by RTMPose. The facial feature-based model considers changes in head position and orientation, while the posture-enhanced model additionally uses upper-body geometric information, including the relative position of the head and shoulder center and shoulder line inclination. Both models include feature interaction terms to capture relationships between gaze features and facial or posture features. Ridge regression is used to learn the residual errors in screen coordinates. The residual is defined as the difference between the actual coordinates for each calibration target and the coordinates predicted by the baseline screen mapping model.

\begin{equation}
    r_i = t_i - \hat{P}_{\theta,i},
    \label{eq:residual}
\end{equation}

where $r_i$ represents $t_i$ the residual,  represents the actual target coordinates, and $\hat{P}_{\theta,i}$ represents the coordinates predicted by the baseline screen mapping model. The trained residual correction model predicts a correction offset using the current gaze features and facial or posture features. However, the reliability of the predicted correction may decrease when the input posture was not observed during calibration or differs significantly from the training samples. To address this limitation, we apply a conditional mechanism that adjusts whether correction is applied and the magnitude of the correction based on the feature ranges of the training data and their similarity to the current input. In addition, the correction offset is limited to a maximum of ±0.15 along each axis in normalized screen coordinates to prevent excessive corrections. When the correction conditions are not satisfied, residual correction is disabled, and the baseline screen mapping result is retained. The final screen coordinates are calculated as follows

\begin{equation}
    \hat{P}_{\mathrm{final}}
    =
    \hat{P}_{c} + \Delta P
    \label{eq:gaze_correction}
\end{equation}

where $\Delta P$ represents the correction offset predicted by the conditional residual correction model. This offset is set to zero when the correction conditions are not satisfied. This mechanism determines whether to apply residual correction based on the relationship between the features collected during calibration and the current input features. It aims to reduce screen coordinate errors caused by posture changes while preventing excessive coordinate shifts when correction is unreliable.

\subsection{Framework }
 The proposed framework consists of five main stages: gaze and pose estimation, calibration, real-time gaze correction, data recording, and visualization. First, the input video from a standard webcam is processed by the gaze estimation model and pose estimation module. For gaze estimation, the system detects the face and uses the ETH-XGaze-based model to estimate gaze direction and extract gaze features for screen coordinate mapping. At the same time, facial position and head orientation features are extracted. The pose estimation module operates asynchronously to detect keypoints around the face and shoulders and constructs posture features representing the spatial relationship between the head and shoulders. When combining the outputs of both modules, the system checks the validity and temporal difference of the pose information. During calibration, the system first checks whether the face and shoulders are visible and establishes a reference posture. The user then looks at nine target points displayed sequentially on the screen. For each target, gaze, facial, and posture features are collected after excluding the initial gaze stabilization period. When the number of samples and their variability satisfy the collection criteria, the system automatically proceeds to the next target. Otherwise, data for the current target are collected again. The collected data are used to train a baseline screen mapping model and residual correction models. The baseline mapping model converts gaze features into user-specific screen coordinates, while the residual correction models learn the differences between the target coordinates and the baseline mapping predictions. Two correction models are compared: one using only facial features and another incorporating posture features. After reviewing the calibration results, the system proceeds to real-time gaze tracking. During real-time tracking, valid gaze features are passed to the baseline mapping model to estimate screen gaze coordinates. A correction offset is applied when the facial and posture features required by the selected correction model are valid and fall within the learned feature range. When RTMPose features are used, only pose information with a timestamp difference of no more than 150 ms from the current gaze frame is incorporated. If the correction conditions are not satisfied, the baseline screen coordinates are used. If gaze estimation itself fails, no new gaze coordinates are generated. During tracking, the system records gaze coordinates and timestamps along with posture features and correction status. Calibration samples, trained mapping and correction parameters, and internal validation results are stored separately in JSON files. For real-time visualization, the estimated gaze position is displayed as a pointer on the screen, while pose keypoints and posture angles are overlaid on the webcam video. This allows users to observe gaze and pose estimation results simultaneously. After tracking ends, the recorded gaze coordinates are used to visualize gaze trajectories and fixation distributions. Gaussian smoothing is applied to generate heatmaps representing gaze paths and spatial dwell distributions.
 
\begin{figure}[h]
  \centering
  \centering
  \includegraphics[width=\linewidth]{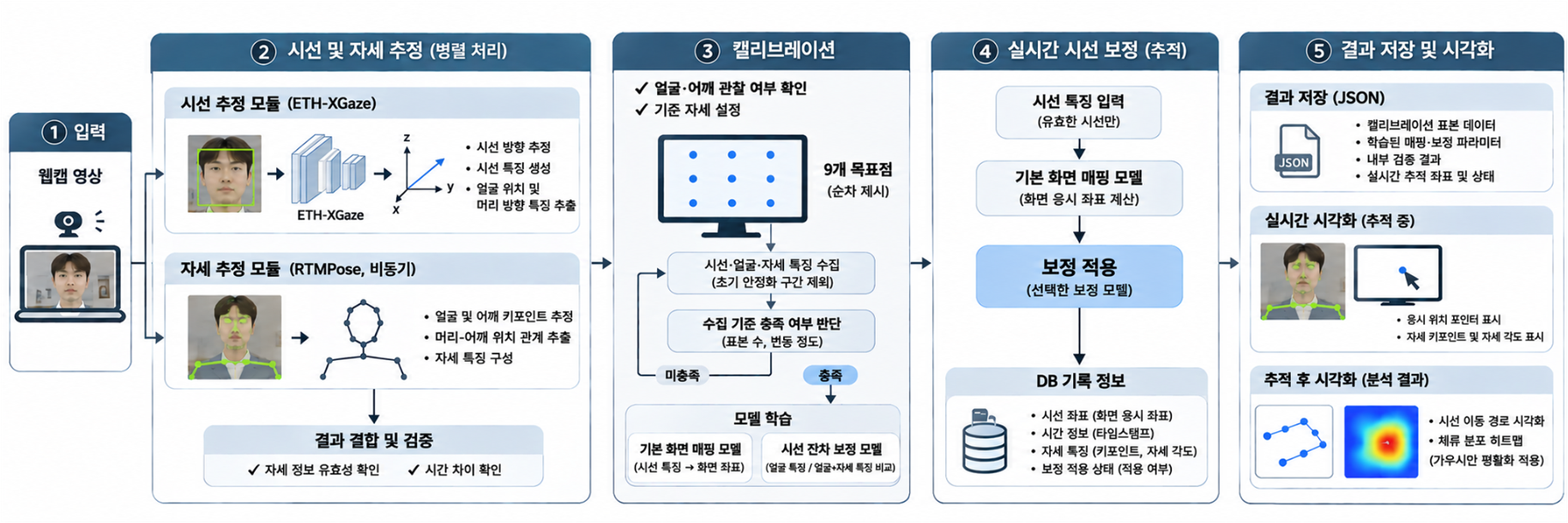}
  \caption{framework}
  \label{fig:rtmposel}
\end{figure}

%% file: sec/4_Experiment.tex
\section{Experiment and Performance Analysis}
\subsection{Date Collenction}
In this study, we used gaze data, posture data, and calibration coordinate data to develop a real-time gaze tracking system using a standard RGB webcam. These data are collected using existing pretrained models and user-specific calibration processes and are used for screen gaze coordinate estimation and conditional residual correction. Gaze data are extracted using XGaze-js, a gaze estimation model based on ETH-XGaze. After preprocessing the detected facial region from the webcam image, a ResNet18-based gaze estimation model estimates the user's horizontal and vertical gaze directions. The estimated gaze directions are converted into two-dimensional gaze features for screen coordinate mapping. Facial position and head orientation features are also collected. These data are used for user-specific screen coordinate mapping and residual correction based on facial characteristics.

\begin{figure}[h]
  \centering
  \centering
  \includegraphics[width=\linewidth]{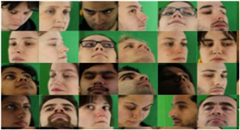}
  \caption{ETH-XGaze Data set}
  \label{fig:rtmposel}
\end{figure}

Posture data are extracted using an RTMPose-based pose estimation model. Keypoints for the nose, eyes, ears, and both shoulders are detected from webcam images. Based on these keypoints, the shoulder center position, distance between the shoulders, relative position of the head and shoulders, and shoulder line inclination are calculated. These measurements are combined into a seven-dimensional posture feature vector and used for conditional residual correction to account for changes in the user's head and upper-body positions. Calibration coordinate data are collected through a user-specific nine-point calibration procedure. The user is instructed to look at nine target points displayed sequentially on the screen. The actual screen coordinates of each target are paired with the gaze features estimated during the corresponding interval. The first second after each target appears is excluded to reduce the effects of gaze transitions. Outliers are then removed from the collected data, and representative gaze features are calculated. The resulting calibration data are used to construct a screen coordinate mapping model based on polynomial ridge regression. In addition, the differences between the target coordinates and the baseline screen mapping predictions are calculated. The conditional residual correction model learns these differences during training.

\begin{figure}[H]
  \centering
  \centering
  \includegraphics[width=\linewidth]{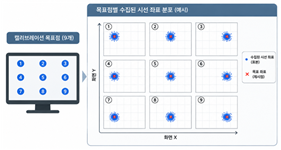}
  \caption{Data Collection Through Calibration}
  \label{fig:rtmposel}
\end{figure}

Finally, the collected gaze, facial, and posture features are paired with their corresponding calibration target coordinates and used for correction model training and real-time gaze tracking. The calibration samples and trained mapping and correction parameters are stored and managed in JSON format.

\subsection{Gaze Tracking Visualization Results}
 Using the developed system, real-time screen gaze coordinates were collected in a standard webcam environment. The system converts gaze directions estimated by XGaze-js into screen coordinates using a user-specific nine-point calibration model. When posture-based correction is enabled, correction offsets derived from RTMPose posture features are applied. After real-time tracking, the collected gaze coordinates are used to generate gaze trajectories and gaze distribution heatmaps. Gaze trajectories represent changes in screen coordinates over time using lines and arrows, allowing the sequence of screen exploration to be visualized. The heatmaps represent the spatial distribution of gaze based on accumulated gaze positions and their durations. In addition, the system identifies major dwell locations and estimates dwell times based on the accumulated time spent within a defined radius.

\begin{table}[t]
    \centering
    \caption{Eye Tracking Result}
    \label{tab:eye_tracking_result}

    \setlength{\tabcolsep}{2pt}
    \renewcommand{\arraystretch}{1.0}

    \begin{tabular}{|c|c|}
        \hline

        \begin{minipage}[c][2.8cm][c]{0.44\columnwidth}
            \centering
            \includegraphics[
                width=\linewidth,
                height=2.3cm,
                keepaspectratio
            ]{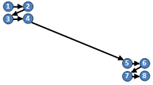}
        \end{minipage}
        &
        \begin{minipage}[c][2.8cm][c]{0.44\columnwidth}
            \centering
            \includegraphics[
                width=\linewidth,
                height=2.3cm,
                keepaspectratio
            ]{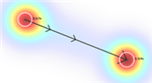}
        \end{minipage}
        \\
        \hline

        \begin{minipage}[c][0.9cm][c]{0.44\columnwidth}
            \centering
            \small Temporal Gaze\\Trajectory
        \end{minipage}
        &
        \begin{minipage}[c][0.9cm][c]{0.44\columnwidth}
            \centering
            \small Gaze Distribution\\and Dwell Locations
        \end{minipage}
        \\
        \hline
    \end{tabular}
\end{table}

\subsection{Experimental Environment}
 In this study, experiments were conducted on a Windows 11 PC equipped with an Intel Core Ultra 5 125H processor and an integrated Neural Processing Unit (NPU) to evaluate the real-time processing performance of the gaze tracking system. Input images were captured using a standard RGB webcam at a resolution of 640 × 480 pixels. Real-time screen gaze coordinates were estimated by combining ETH-XGaze-based gaze direction estimation with RTMPose-based pose estimation.

\subsection{Accuracy Comparison with Existing Systems and Conditional Correction Analysis}

To evaluate the gaze estimation performance of the proposed system, we selected EyeGestures~[20], gaze\_tracker~[21], and xgaze-js as baseline methods. All experiments were conducted using a standard RGB webcam, with a single participant evaluating each system. After completing the calibration procedure for each system, the participant was instructed to look at 12 target points sequentially displayed on the screen. Each target was presented for 3 seconds. To reduce the effects of gaze transitions, only valid gaze coordinates obtained after excluding the first second of each target presentation were analyzed. The evaluation metric was defined as the Euclidean distance between the center of each target and the estimated gaze coordinates. All coordinates were converted to a common resolution of $1920 \times 1080$ pixels, matching the resolution of the original target image. The mean error was calculated for each target, and the overall mean error was obtained by assigning equal weight to all 12 targets.


\begin{figure}[H]
    \centering
    \includegraphics[width=\linewidth]{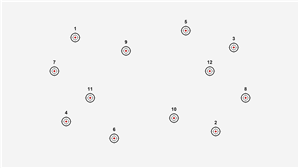}
    \caption{Target Image Used in the Experiment}
    \label{fig:rtmposel}
\end{figure}

\begin{table}[H]
\centering
\caption{Experimental Results (Visualization)}
\label{tab:visualization_1}
\renewcommand{\arraystretch}{1.2}
\setlength{\tabcolsep}{3pt}

\begin{tabular}{|c|c|}
\hline
\textbf{Proposed System} &
\textbf{EyeGestures} \\
\hline
\includegraphics[width=0.43\linewidth]{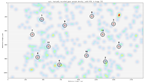} &
\includegraphics[width=0.43\linewidth]{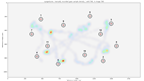} \\
\hline
\end{tabular}
\end{table}

\begin{table}[H]
\centering
\caption{Experimental Results (Visualization)}
\label{tab:visualization_2}
\renewcommand{\arraystretch}{1.2}
\setlength{\tabcolsep}{3pt}

\begin{tabular}{|c|c|}
\hline
\textbf{gaze\_tracker} &
\textbf{xgaze} \\
\hline
\includegraphics[width=0.43\linewidth]{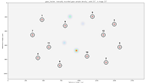} &
\includegraphics[width=0.43\linewidth]{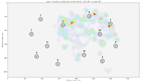} \\
\hline
\end{tabular}
\end{table}

\begin{table}[H]
\centering
\caption{Experimental Results (Quantitative)}
\label{tab:quantitative_results}
\renewcommand{\arraystretch}{1.3}
\setlength{\tabcolsep}{4pt}
\small

\begin{tabular}{|l|c|c|c|}
\hline
\textbf{Compared} &
\textbf{Targets with} &
\textbf{Mean} &
\textbf{Valid} \\
\textbf{System} &
\textbf{Valid Outputs} &
\textbf{Distance} &
\textbf{Output} \\
&
&
\textbf{Error (px)} &
\textbf{Rate (\%)} \\
\hline
Proposed System & 12/12 & 204.24 & 100.00 \\
\hline
EyeGestures & 9/12 & 191.49$^{*}$ & 70.17 \\
\hline
gaze\_tracker & 12/12 & 482.68 & 96.74 \\
\hline
xgaze-js & 12/12 & 318.30 & 100.00 \\
\hline
\end{tabular}
\end{table}

The experimental results show that the proposed system achieved a mean screen coordinate error of 204.24 pixels, lower than those of xgaze-js (318.30 pixels) and gaze\_tracker (482.68 pixels). The proposed system produced valid outputs for all 12 targets, with a valid output rate of 100\% during the evaluation intervals. In contrast, EyeGestures produced valid outputs for 9 of the 12 targets. For these 9 commonly valid targets, EyeGestures achieved a mean error of 191.49 pixels, slightly lower than the 196.91 pixels obtained by the proposed system. An analysis of 504 frames recorded during real-time tracking showed that residual correction was applied to 417 frames (82.74\%). Specifically, facial feature-based correction was applied to 343 frames (68.06\%), while combined facial and posture feature-based correction was applied to 74 frames (14.68\%). For the remaining 87 frames (17.26\%), the correction conditions were not satisfied, and the baseline screen mapping predictions were used. The 95th percentile of the error across valid frames was 744.60 pixels, indicating that large estimation errors occurred in some frames. This study presents a preliminary comparative experiment involving a single participant. Furthermore, the individual contributions of the baseline screen mapping model and the residual correction models to estimation accuracy were not independently evaluated. Therefore, further validation is needed to assess generalization across different users and posture conditions, as well as the effectiveness of individual correction components.

%% file: sec/5_conclusion.tex
\section{Conclusion and Future Work}
\subsection{Conclusion and Limitation}
In this study, we developed a real-time gaze tracking system that estimates screen gaze coordinates using a standard RGB webcam. To achieve this, we combined an ETH-XGaze-based gaze estimation model with user-specific calibration and RTMPose-based posture estimation. We also developed a conditional residual correction method that estimates screen coordinate residuals using facial and posture features. This method determines whether to apply correction based on the validity of the input features and the distribution of the training data. To evaluate the proposed system, we conducted a preliminary experiment with a single participant using 12 target points displayed on the screen. We compared the proposed system with EyeGestures, gaze\_tracker, and XGaze-js, which served as the base model for our system. All comparison systems use standard RGB webcams for gaze tracking. The experimental results showed that the proposed system achieved a mean screen coordinate error of 204.24 pixels, which was lower than those of XGaze-js and gaze\_tracker. The system also produced valid gaze coordinates for all target points, confirming its ability to perform real-time gaze tracking. However, the proposed system did not consistently outperform EyeGestures when evaluated on the target points shared by both systems. Relatively large estimation errors were also observed in some frames. These results demonstrate the feasibility and preliminary performance of the proposed system in a standard RGB webcam environment. However, the study has limitations related to the experimental scale and evaluation conditions. First, the experiment involved only one participant, making it difficult to examine performance differences caused by individual facial characteristics, gaze movements, and posture changes. In addition, the comparison systems differed in their gaze estimation methods, calibration processes, and recommended camera settings. These differences made it difficult to compare their performance under identical conditions. Therefore, the current results are not sufficient to establish the overall superiority of the proposed system. Further evaluation with more participants and different experimental conditions is needed.

\subsection{Implications}
With advances in artificial intelligence, gaze tracking technology is being applied in a growing number of fields. According to a market research report by Fortune Business Insights, the gaze tracking market is expected to continue growing. However, the cost of dedicated equipment, performance degradation caused by changes in users and environments, and difficulties with calibration remain barriers to the wider adoption of this technology [22]. In this context, our study developed a real-time screen gaze tracking system using a standard RGB webcam. The study examined the possibility of applying gaze tracking technology without dedicated eye tracking equipment. Previous studies have explored various approaches, including gaze direction estimation, gaze region classification, and gaze prediction based on body movements. In contrast, our study converts gaze estimates from an ETH-XGaze-based model into two-dimensional screen coordinates through user-specific calibration, allowing the system to adapt to different display environments. We also incorporated posture features extracted by RTMPose into the residual correction of screen coordinates. This approach goes beyond combining gaze estimation with posture analysis by introducing a screen coordinate estimation framework that also considers the conditions under which correction is applied. From a practical perspective, using a standard RGB webcam may reduce the need for dedicated eye tracking equipment. This could help extend the use of gaze tracking technology beyond specialized hardware environments to everyday computer settings. Furthermore, the proposed approach suggests the potential for future applications of real-time gaze tracking systems in mobile and on-device environments, as well as conventional computer systems.

\subsection{Future Work}
In future work, we plan to expand our evaluation to include more participants and different camera conditions. We will also compare baseline screen coordinate mapping, facial feature-based residual correction, and residual correction combining facial and posture features in stages to quantitatively evaluate the contribution of each component. In addition, we plan to integrate a 3D embedding method into the RTMPose-based posture estimation framework. This will allow us to examine the relative position and distance changes between the user and the webcam, as well as head and upper-body posture, in greater detail. Through this approach, we aim to incorporate three-dimensional spatial features into the gaze coordinate correction process, extending the existing correction method based on two-dimensional posture features. Our goal is to reduce gaze estimation errors caused by changes in user-to-camera distance and posture, and to improve the accuracy and robustness of the real-time gaze tracking system.